\documentclass[11pt,a4paper]{article}
\usepackage[hyperref]{acl2020}
\usepackage{times}
\usepackage{latexsym}

\usepackage{microtype}

\usepackage{amsmath}
\usepackage{amsfonts}
\usepackage{algorithm}
\usepackage{algorithmic}

\usepackage{float}
\usepackage{multirow}
\usepackage{verbatim}
\usepackage{caption}
\usepackage{subcaption}
\usepackage{graphicx}

\usepackage{xcolor}

\newcommand{\RN}[1]{%
	\textup{\uppercase\expandafter{\romannumeral#1}}%
}

\aclfinalcopy % Uncomment this line for the final submission
\title{Salience Estimation with Multi-Attention Learning for Abstractive Text Summarization}

	\author{
		Piji Li$^{\dag\ddag}$  \ \ Lidong Bing$^{\natural}$  \ \ Zhongyu Wei$^{\S}$ \ \ Wai Lam$^{\dag}$\\
		$^{\dag}$Department of Systems Engineering and Engineering Management,\\
		The Chinese University of Hong Kong\\
		$^{\ddag}$Tencent AI Lab, Shenzhen, China\\
		$^{\natural}$Alibaba DAMO Academy\\
		$^{\S} $School of Data Science, Fudan University, China\\
		{\tt  $^{\dag}$\{pjli, wlam\}@se.cuhk.edu.hk}\\
		{\tt $^{\natural}$l.bing@alibaba-inc.com, $^{\S}$zywei@fudan.edu.cn}
	}

\date{}

\begin{document}
\maketitle
\begin{abstract}
Attention mechanism plays a dominant role in the sequence generation models and has been used to improve the performance of machine translation and abstractive text summarization. 
Different from neural machine translation, in the task of text summarization, salience estimation for words, phrases or sentences is a critical component, since the output summary is a distillation of the input text.   
Although the typical attention mechanism can conduct text fragment selection from the input text conditioned on the decoder states, there is still a gap to conduct direct and effective salience detection. To bring back direct salience estimation for summarization with neural networks, we propose a Multi-Attention Learning framework which contains two new attention learning components for salience estimation: supervised attention learning and unsupervised attention learning. We regard the attention weights as the salience information, which means that the semantic units with large attention value will be more important. 
The context information obtained based on the estimated salience is incorporated with the typical attention mechanism in the decoder to conduct summary generation. 
Extensive experiments on some benchmark datasets in different languages demonstrate the effectiveness of the proposed framework for the task of abstractive summarization.
\end{abstract}
\section{Introduction}
	\label{sec:intro}
	
	Sequence-to-sequence (seq2seq) framework with attention mechanism has achieved significant improvement in the field of neural machine translation \cite{bahdanau2014neural}.
	Encouraged by this outcome, some researchers transplanted the seq2seq framework to tackle the problem of abstractive text summarization \cite{rush2015neural,chopra2016abstractive,nallapati2016abstractive} and also obtained some encouraging results. Since then, abstractive text summarization has bloomed into a popular research task and quite a few seq2seq-based frameworks have been proposed. 
	For example, \citet{see2017get} integrated the copy operation \cite{gu2016incorporating,vinyals2015pointer} and the coverage model \cite{tu2016modeling} into the typical attention based seq2seq to generate better summaries. \citet{li2017deep} designed a recurrent generative decoder to capture the latent structures in the target summaries. \citet{paulus2017deep} employed deep reinforcement learning techniques to enhance the performance of this task.
	
	The above frameworks can improve the quality of the generated abstractive summaries to some extent. However, when we immerse ourselves in designing such dazzling and complex tricks on top of the seq2seq model, we may unintentionally ignore some important characteristics specific to the task of text summarization.
	Along the whole way of summarization research, salience detection---finding the most important information (words, phrases, or sentences) from the source input text---has always been the most crucial and essential component. Some supervised \cite{ng2012exploiting,wang2013sentence} or unsupervised \cite{erkan2004lexrank,mihalcea2004textrank} learning methods were proposed to estimate the salience score for producing better summaries.  
	However, for the attention-based seq2seq framework, it is not straightforward to figure out how to conduct salience detection. 
	%We are sure that the attention mechanism plays an important role in detecting the salient fragments from the source input text. 
	The current attention mechanism for the summarization task is not as natural and effective as in some other tasks. For instance, in neural machine translation, it is reasonable to use the current decoding state to attend the source sequence to get the relevant information for translating the next target word. In reading comprehension, it makes sense to use the question to attend the reading passage to retrieve relevant information for extracting the answer.
	But for text summarization, it is difficult to connect the attention mechanism with the salience estimation operation. 
	Although several works have tried some strategies to conduct the salience detection, there still exist some limitations. For example, the selective mechanism \cite{zhou2017selective} only implicitly performs salience detection. The graph-based attention mechanism \cite{tan2017abstractive} only adopts an unsupervised method, thus it is not capable to exploit the supervised signal in the training data. 
	
	In this paper, we propose two global attention mechanisms based on supervised learning and unsupervised learning respectively for salient information detection. For the supervised attention mechanism, we employ a supervised learning method to estimate the probability of each word in the input text to be included in the generated summary. The normalized probability value is regarded as the supervised attention signal. For the unsupervised attention mechanism, inspired by the PageRank \cite{page1999pagerank} based text summarization methods such as LexRank \cite{erkan2004lexrank} and TextRank \cite{mihalcea2004textrank}, as well as the graph-based attention mechanism \cite{tan2017abstractive}, we employ the PageRank algorithm to estimate the salience score of each input word, which is regarded as the unsupervised attention signal. Thus, these two types of attention signals contain the salience information of the terms in the source text. 
	To examine the efficacy of the obtained salience information, we integrate these signals into a simple base model for abstractive summarization, i.e. the attention based seq2seq model. Note that we do not employ more sophisticated and powerful models, because the aim of this work is to verify that bringing back salience estimation for neural abstractive summarization is helpful to improve the performance, where a simple base model allows the conclusion not biased by other modeling structures.
	
	Our main contributions are summarized as follows.
	(1) We investigate a crucial element of text summarization problem, namely salience estimation, which has been overlooked by the prior neural abstractive summarization approaches. 
	(2) We propose a supervised attention mechanism to directly estimate the salience under the supervision signal provided by the state of the input text, and an unsupervised attention mechanism which employs a graph algorithm to estimate the salience of each input word. 
	(3) We integrate the two types of attention information into a base model and propose a unified neural network based framework, named Multi-Attention Learning (MAL), to tackle the task of abstractive summarization.
	(4) Experimental results on some benchmark datasets in different languages demonstrate the effectiveness of the proposed attention learning methods for salience estimation.

	\section{Our Framework}
	
	\subsection{Overview}
	
	\begin{figure*}[!t]
		\centering
		\includegraphics[width=1.8\columnwidth]{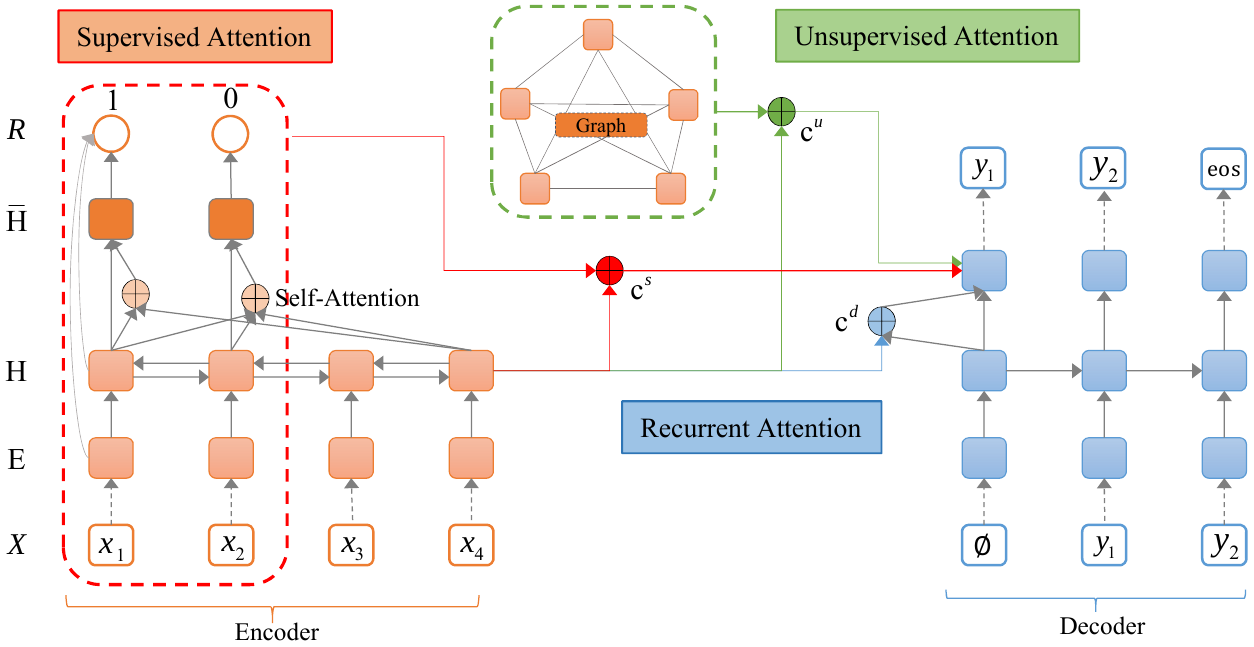}
		\caption{Our Multi-Attention Learning (MAL) framework for abstractive summarization.}
		\label{fig:framework}
	\end{figure*}
	
	The proposed Multi-Attention Learning (MAL) framework is shown in Figure~\ref{fig:framework}. The input is a variable-length sequence ${X} = (x_1, x_2, \ldots, x_m)$, representing the source text.
	The output ground truth is also a sequence ${Y} = (y_1, y_2, \ldots, y_n)$.
	We denote the generated summary sequence as $\hat{Y} = (\hat{y}_1, \hat{y}_2, \ldots, \hat{y}_{n'})$.
	%The basic structure of MAL framework is a seq2seq model with attention mechanism \cite{bahdanau2014neural}.
	For global salience estimation, we add two tailor-made attention learning mechanisms: supervised attention learning and unsupervised attention learning.
	The aim of supervised attention learning is to predict if words from the input source text should be selected into the generated summaries, i.e. predicting a $0$ or $1$ label for each word. As shown in Figure~\ref{fig:framework}, the word embeddings $\mathbf{E}$ and the encoder  recurrent neural network (RNN) hidden states $\mathbf{H}$ are taken as the input information of this supervised attention learning modular. We also design a self-attention model to capture more context information from the source text for better feature representation learning. The output of this component is regarded as the supervised attention information $\mathbf{a}^s$.
	For unsupervised attention learning, we employ the PageRank algorithm to estimate the salience score of each input words in an unsupervised manner. We treat the salience score as the unsupervised attention information $\mathbf{a}^u$. Then these two types of global attention information representing word salience are combined with the hidden states $\mathbf{H}$ of the input source text to obtain the global attention context. Finally, the attention context information is incorporated in the decoding procedure to generate the abstractive summaries.

	\subsection{Supervised Attention Learning}
	\label{sec:sal}
	
	The aim of supervised attention learning is to estimate the probability of the words in the source text to appear in the generated summaries. With sufficient training data, we can regard this problem as a supervised sequence labeling task.
	We employ a straightforward method to prepare the ground truth labels $\mathbf{R}$ for the source text.
	The words in the source text (except the stopwords) that appear in the ground truth summaries are annotated with the positive label $1$. All other words and the punctuations are annotated with the negative label $0$.

	The structure of the supervised attention learning framework (we illustrate the computational logic with the first two states) is depicted on the left of Figure~\ref{fig:framework}. We first map each input word $x_t$ into a vector $\mathbf{x}_t \in \mathbb{R}^{k_e}$ by retrieving an embedding lookup table, which is  randomly initialized and fine-tuned in the training procedure.
	The word embedding sequence is fed into a bi-directional RNN to capture the context information.
	Compared with LSTM \cite{hochreiter1997long}, GRU \cite{cho2014learning} has comparable performance but with less parameters and more efficient computation, so we employ GRU as the basic recurrent unit:
	\begin{equation}
	\begin{array}{l}
	\mathbf{r}_t = \sigma (\mathbf{W}_{xr}\mathbf{x}_t + \mathbf{W}_{hr}\mathbf{h}_{t - 1} + \mathbf{b}_r)\\
	\mathbf{z}_t = \sigma (\mathbf{W}_{xz}\mathbf{x}_t + \mathbf{W}_{hz}\mathbf{h}_{t - 1} + \mathbf{b}_z)\\
	\mathbf{g}_t = \tanh (\mathbf{W}_{xh}\mathbf{x}_t + \mathbf{W}_{hh}(\mathbf{r}_t \odot \mathbf{h}_{t - 1}) + \mathbf{b}_h)\\
	\mathbf{h}_t = \mathbf{z}_t \odot \mathbf{h}_{t - 1} + (1 - \mathbf{z}_t) \odot \mathbf{g}_t
	\end{array}
	\label{eq:gru}
	\end{equation}
	where $\mathbf{r}_t$ is the reset gate, $\mathbf{z}_t$ is the update gate to control the mixture of the previous hidden $\mathbf{h}_{t - 1}$ and $\mathbf{g}_t$ to get the current hidden $\mathbf{h}_t\in \mathbb{R}^{k_h}$, and those $\mathbf{W}$'s and $\mathbf{b}$'s are learnable parameters.
	$\odot$ denotes the element-wise multiplication, and $tanh$ is the  hyperbolic tangent activation function.
	We employ a bidirectional GRU network to produce two hidden states at the time step $t$:
	\begin{equation}
	\begin{array}{l}
	{{\mathord{\buildrel{\lower3pt\hbox{$\scriptscriptstyle\rightharpoonup$}} 
				\over {\mathbf{h}}} }_t} = GRU({\mathbf{x}_t},{{\mathord{\buildrel{\lower3pt\hbox{$\scriptscriptstyle\rightharpoonup$}} 
				\over {\mathbf{h}}} }_{t - 1}})\\
	{{\mathord{\buildrel{\lower3pt\hbox{$\scriptscriptstyle\leftharpoonup$}} 
				\over {\mathbf{h}}} }_t} = GRU({\mathbf{x}_t},{{\mathord{\buildrel{\lower3pt\hbox{$\scriptscriptstyle\leftharpoonup$}} 
				\over {\mathbf{h}}} }_{t + 1}})
	\end{array}
	\end{equation}
	Then the overall hidden state $\mathbf{h}_t^e \in \mathbb{R}^{2k_h}$ of the encoder is a concatenation of both directions: 
	\begin{equation}
	{\mathbf{h}}_t^e = {{\mathord{\buildrel{\lower3pt\hbox{$\scriptscriptstyle\rightharpoonup$}} 
				\over {\mathbf{h}}} }_t}||\mathord{\buildrel{\lower3pt\hbox{$\scriptscriptstyle\leftharpoonup$}} 
		\over {\mathbf{h}}}_t 
	\end{equation}
	
	In order to capture more context information of the input sequence, we integrate a Self-Attention modeling component. The self-attention weight at the time step $t$ is calculated based on the relationship of $\mathbf{h}_t^{e}$ and all the source hidden states $\{\mathbf{h}_i^e\}$. Let $a^e_{i,j}$ be the attention weight between $\mathbf{h}_i^{e}$ and $\mathbf{h}_j^{e}$ which can be calculated as:
	\begin{equation}
	\begin{aligned}
	{a^e_{i,j}} &= \frac{{\exp ({e_{i,j}})}}{{\sum\nolimits_{j' = 1}^{{T^e}} {\exp ({e_{i,j'}})} }}\\
	{e_{i,j}} &= {\mathbf{v}^\top_e}\tanh (\mathbf{W}_{hi}^e\mathbf{h}_i^{{e}} + \mathbf{W}_{hj}^e\mathbf{h}_j^e + {\mathbf{b}^e_a})
	\end{aligned}
	\end{equation}
	where $\mathbf{W}_{hi}^e \in \mathbb{R}^{k_h \times 2k_h}$, $\mathbf{W}_{hj}^e \in \mathbb{R}^{k_h \times 2k_h}$, $\mathbf{b}^e_a \in \mathbb{R}^{k_h}$, and $\mathbf{v}_e \in \mathbb{R}^{k_h}$.
	Then the self-attention context is obtained by the weighted linear combination of all the source hidden states:
	\begin{equation}
	{\mathbf{c}^e_t} = \sum\nolimits_{j' = 1}^{{T^e}} {{a^e_{t,j'}}\mathbf{h}_{j'}^e} 
	\end{equation}
	where $T^e$ is the sequence length. The original hidden state $\mathbf{h}^e_t$ can be revised using the self-attention context information $\mathbf{c}^e_t$:
	\begin{equation}
	\bar{\mathbf{h}}^e_t = \tanh (\mathbf{W}_{h\bar{h}}^e\mathbf{h}_t^{e} + \mathbf{W}_{c\bar{h}}^e\mathbf{c}_t^e + {\mathbf{b}^e_{\bar{h}}})
	\end{equation}
	
	Finally, as shown in Figure~\ref{fig:framework}, we feed the word embedding vector $\mathbf{x}_t$, the hidden state $\mathbf{h}^e_t$, the self-attention context $\mathbf{c}^e_t$, and the self-attention state $\bar{\mathbf{h}}^e_t$ into the final output layer to get the prediction $\hat r \in (0, 1)$:
	\begin{equation}
	\hat r = \sigma(\mathbf{W}_{xr}\mathbf{x}_t + \mathbf{W}_{hr}\mathbf{h}_t^{e} + \mathbf{W}_{cr}\mathbf{c}_t^e + \mathbf{W}_{\bar{h}r}\bar{\mathbf{h}}_t^e + {\mathbf{b}^e_r})
	\end{equation}
	where $\sigma()$ is the sigmoid function. The value of $\hat r$ represents the salience of the corresponding words in the source text.
	
	In order to get the attention information and the attention context information, we first add a normalization procedure to the predicted $ \mathbf{\hat r}$:
	\begin{equation}
	a^s_i = \frac{{\exp ({\hat r_{i}})}}{{\sum\nolimits_{j' = 1}^{{T^e}} {\exp ({\hat r_{j'}})} }}
	\end{equation} 
	We regard the vector $\mathbf{a}^s \in \mathbb{R}^{T^e}$ as the supervised attention information.
	
	Based on the supervised attention information $\mathbf{a}^s$, we can obtain one type of global attention context by the weighted linear combination of the source hidden states:
	\begin{equation}
	{\mathbf{c}^s} = \sum\nolimits_{j' = 1}^{{T^e}} {{a^s_{j'}}\mathbf{h}_{j'}^e} 
	\end{equation}
	Finally, $\mathbf{c}^s$ is incorporated in the decoder as the supervised attention context information for the summary generation.
	
	\subsection{Unsupervised Attention Learning}
	\label{sec:ual}
	
	In the traditional text summarization research, the PageRank \cite{page1999pagerank} based salience estimation methods play a crucial role in identifying the most important information from the source text. Some classical methods such as LexRank \cite{erkan2004lexrank} and TextRank \cite{mihalcea2004textrank} were proposed to tackle the problems of text summarization and keyphrase extraction, and have been applied into practical summarization applications and products. 
	%Unfortunately, there are very few works realized this problem and consider it when conduct the research in neural abstractive summarization.
	\citet{tan2017abstractive} introduced the graph-based attention mechanism into the seq2seq framework for sentence salience estimation and obtained encouraging results. Here, we also employ PageRank algorithm to conduct the unsupervised attention learning for salience estimation, as depicted in the middle-upper part of Figure~\ref{fig:framework}. The difference is that we conduct the learning on word level to estimate the salience.
	
	For an input text sequence $X$ with length $m$, and $\mathbf{x}_t \in \mathbb{R}^{k_e}$ representing the embedding vector for the word $x_t$, to build the word-based graph $G$, we take the nonstop words as the vertex set $V$, and the relations between the words, computed with Equation \ref{eq:edge}, as edge set $E$. We employ a parameterized tensor method to calculate the weights of the edges. Assume that the adjacent matrix is $\mathbf{M} \in \mathbb{R}^{m \times m}$, then each element can be calculated as: 
	\begin{equation}
	\label{eq:edge}
	\mathbf{M}_{i,j} = \mathbf{x}_i^\top\mathbf{W}^p\mathbf{x}_j
	\end{equation}
	where $\mathbf{W}^p \in \mathbb{R}^{k_e \times k_e}$ is a neural parameter to be learned.  PageRank is a iterative algorithm, but we can get the closed form as discussed in \cite{tan2017abstractive}:
	\begin{equation}
	\mathbf{p} = (1 - d)(\mathbf{I} - d\mathbf{M}\mathbf{D}^{-1})^{-1}\mathbf{q} 
	\end{equation}
	where $\mathbf{D}$ is a diagonal matrix and $\mathbf{D}(i,i) = \sum {\mathbf{M}(:,i)}$, $d$ is the damping factor, and $\mathbf{q} \in \mathbb{R}^m$ with all the elements equal to $1/m$.
	Then the vector $\mathbf{p} \in \mathbb{R}^m$ is the estimated salience score for all the $m$ words.
	We also add a normalization procedure to $\mathbf{p}$:
	\begin{equation}
	a^u_i = \frac{{\exp ({p_{i}})}}{{\sum\nolimits_{j' = 1}^{{m}} {\exp ({p_{j'}})} }}
	\end{equation} 
	Then the vector $\mathbf{a}^u \in \mathbb{R}^{m}$ is regarded as the unsupervised attention information.
	We can also obtain the second type of global attention context by the weighted linear combination of the word embeddings using $\mathbf{a}^u$:
	\begin{equation}
	{\mathbf{c}^u} = \sum\nolimits_{j' = 1}^{{m}} {{a^u_{j'}}\mathbf{x}_{j'}} 
	\end{equation}
	And $\mathbf{c}^u$ will be incorporated with the seq2seq framework as the unsupervised attention context information.
	
	\subsection{Summary Generation}
	\label{sec:generation}
	
	The decoder of our MAL framework is still a GRU based recurrent neural network with improved attention modeling. The first hidden state $\mathbf{h}_1^d$ of the decoder is initialized using the average of all the source input hidden states: 
	$
	\mathbf{h}_1^d = \frac{1}{{{m}}}\sum\limits_{t = 1}^{{m}} {\mathbf{h}_t^e}
	$.
	Then the two layers of GRUs are designed to conduct the attention weights calculation and decoder hidden states update.
	On the first GRU layer, the hidden state is calculated only using the current input word embedding $\mathbf{y}_{t-1}$ and the  previous hidden state $\mathbf{h}_{t-1}^{d_1}$:
	\begin{equation}
	\mathbf{h}_t^{d_1} = GRU_1(\mathbf{y}_{t-1}, \mathbf{h}_{t-1}^{d_1})
	\end{equation}
	Then the attention weights at the time step $t$ are calculated based on the relationship of $\mathbf{h}_t^{d_1}$ and all the source hidden states $\{\mathbf{h}_t^e\}$:
	\begin{equation}
	\begin{aligned}
	{a_{i,j}^d} &= \frac{{\exp ({e_{i,j}})}}{{\sum\nolimits_{j' = 1}^{{T^e}} {\exp ({e_{i,j'}})} }}\\
	{e_{i,j}} &= {\mathbf{v}^\top}\tanh (\mathbf{W}_{hh}^d\mathbf{h}_i^{{d_1}} + \mathbf{W}_{hh}^e\mathbf{h}_j^e + {\mathbf{b}_a})
	\end{aligned}
	\end{equation}
	The attention context is obtained by the weighted linear combination of all the source hidden states:
	$
	{\mathbf{c}_t^d} = \sum\nolimits_{j' = 1}^{{T^e}} {{a_{t,j'}^d}\mathbf{h}_{j'}^e} 
	$. The final hidden state $\mathbf{h}_t^{d_2}$ is the output of the second GRU layer, jointly considering the word $\mathbf{y}_{t-1}$, the previous hidden state $\mathbf{h}_{t-1}^{d_2}$, and the attention context $\mathbf{c}_t^d$:
	\begin{equation}
	\mathbf{h}_t^{d_2} = GRU_2(\mathbf{y}_{t-1}, \mathbf{h}_{t-1}^{d_2}, \mathbf{c}_t^d)
	\end{equation}
	The traditional seq2seq framework will predict the target word based on $\mathbf{h}_t^{d_2}$.
	
	\subsubsection{Multi-Attention Integration}
	
	Recall that we have obtained the supervised attention context $\mathbf{c}^s$ in Section~\ref{sec:sal} and the unsupervised attention context $\mathbf{c}^u$ in Section~\ref{sec:ual}.  Then we integrate all the attention context information here in a straightforward manner:
	\begin{equation}
	\mathbf{h}_t^a = \tanh({\mathbf{W}_{ha}^{d}}\mathbf{h}_{t}^{d_2} + \mathbf{W}_{c^sa}^d\mathbf{c}^s +  \mathbf{W}_{c^ua}^d\mathbf{c}^u  + {\mathbf{b}^{d}_{ha}})
	\end{equation}
	Finally, the probability of generating any target word $y_t$ is given as follows:
	\begin{equation}
	\hat{\mathbf{y}}_t = \varsigma({\mathbf{W}_{hy}^{d}}\mathbf{h}_{t}^{a} + {\mathbf{b}^{d}_{hy}})
	\end{equation}
	where ${\mathbf{W}_{hy}^{d}} \in \mathbb{R}^{k_y \times k_h}$ and ${\mathbf{b}^{d}_{hy}} \in \mathbb{R}^{k_y}$. $\varsigma(\cdot)$ is the softmax function.
	In the prediction state, we use the beam search algorithm \cite{koehn2004pharaoh} for decoding and generating the best summary.
	
	\subsection{Model Training}
	For supervised attention learning,  we use the cross-entropy as the objective function which need to be minimized:
	\begin{equation}
	\mathcal{L}_s = -\frac{1}{m}\sum\limits_i {{r_i}\log ({{\hat r}_i}) + (1 - {r_i})\log (1 - {{\hat r}_i})}
	\end{equation}
	where $\hat r_i$ and $r_i$ is the prediction and the ground truth respectively.
	
	For summary generation, we employ the negative log likelihood (NLL) as the objective function. Given the ground truth summary
	${Y} = \{\mathbf{y}_1, \mathbf{y}_2, \ldots, \mathbf{y}_n\}$ for the input sequence $X$, we have:
	\begin{equation}
	\mathcal{L}_{NLL} = -\sum\limits_{t = 1}^n {\log p({y_t}|{y_{<t}},X)}
	\end{equation}
	Then the final objective loss function is:
	\begin{equation}
	\mathcal{L} = \mathcal{L}_s + \mathcal{L}_{NLL}
	\end{equation}
	The whole framework can be trained using the multi-task learning paradigm with the back-propagation method in an end-to-end training style. 
	Adadelta \cite{zeiler2012adadelta} with hyperparameters $\rho = 0.95$ and $\epsilon = 1e-6$ is used for gradient based optimization.  
	
	\section{Experimental Setup}
	
	\subsection{Datasets}
	We train and evaluate our framework on three popular benchmark datasets.
	\textbf{Gigawords} is an English sentence summarization dataset prepared based on Annotated Gigawords\footnote{https://catalog.ldc.upenn.edu/ldc2012t21} by extracting the first sentence from a news report together with the headline to form a source and summary pair (i.e. the first sentence and headline).
	We directly download the prepared dataset used in \cite{rush2015neural}.
	It roughly contains 3.8M training pairs, 190K validation pairs, and 2,000 test pairs. The test set is identical to the one used in all the comparative methods.
	\textbf{DUC-2004}\footnote{http://duc.nist.gov/duc2004} is another English dataset only used for testing, where we directly apply the model trained from Gigawords. It contains 500 documents. Each document contains 4 model summaries written by experts. The length of the summary is limited to 75 bytes.
	\textbf{LCSTS} is a large-scale Chinese short text summarization dataset, consisting of pairs of short text and summary, collected from Sina Weibo\footnote{http://www.weibo.com} \cite{hu2015lcsts}.
	We take Part-I as the training set, Part-II as the development set, and Part-III as the test set. There is a score in the range of $1\sim5$ labeled by human to indicate the relevance between an article and its summary. We only make use of those pairs with scores no less than 3. Thus, the three parts contain 2.4M, 8.7k, and 725 data points respectively.
	In our experiments, we directly take the Chinese character sequences as input, without performing word segmentation.

	\subsection{Evaluation Metrics}
	We use ROUGE \cite{lin2004rouge} with standard options as our evaluation metric.
	The idea of ROUGE is to count the number of overlapping units between the generated summaries and the reference summaries, such as overlapped n-grams, word sequences, and word pairs.
	F-measures of ROUGE-1 (R-1), ROUGE-2 (R-2) and ROUGE-L (R-L) are reported for Gigawords and LCSTS datasets. ROUGE recalls are reported for the DUC dataset.
	
	\subsection{Comparative Methods}
	We compare our \textbf{MAL} with a bunch of previous methods.
	Since the datasets are quite standard, so we just extract the results from their papers, if reported. Therefore, the compared methods on different datasets may be slightly different.
	\textbf{TOPIARY} \cite{zajic2004bbn} is the best on DUC2004 Task-1 for compressive text summarization.	It combines a system using linguistic based transformations and an unsupervised topic detection algorithm for compressive text summarization.
	\textbf{MOSES+} \cite{rush2015neural} uses a phrase-based statistical machine translation system trained on Gigaword to produce summaries.
	%It also augments the phrase table with ``deletion'' rules to improve the baseline performance, and MERT is also used to improve the quality of generated summaries.
	\textbf{ABS} and \textbf{ABS+} \cite{rush2015neural} are both the neural network based models with local attention modeling for abstractive sentence summarization.
	%ABS+ is trained on the Gigaword corpus, but combined with an additional log-linear extractive summarization model with handcrafted features.
	\textbf{RNN} and \textbf{RNN-context} \cite{hu2015lcsts} are two seq2seq architectures. RNN-context integrates attention mechanism to model the context.
	\textbf{CopyNet} \cite{gu2016incorporating} integrates a copying mechanism into the seq2seq framework.
	\textbf{RNN-distract} \cite{chen2016distraction} uses a new attention mechanism by distracting the historical attention in the decoding steps.
	\textbf{RAS-LSTM} and \textbf{RAS-Elman} \cite{chopra2016abstractive} both consider words and word positions as input and use convolutional encoders to handle the source information.
	For the attention based sequence decoding process, RAS-Elman selects Elman RNN \cite{elman1990finding} as its decoder, and RAS-LSTM selects LSTM architecture \cite{hochreiter1997long}.
	\textbf{LenEmb} \cite{kikuchi2016controlling} uses a mechanism to control the summary length by considering the length embedding vector as the input.
	\textbf{ASC+FSC$_1$} \cite{miao2016language} uses a generative model with attention mechanism to tackle the sentence compression problem.
	%The model first draws a latent summary sentence from a background language model, and then subsequently draws the observed sentence conditioned on this latent summary.
	\textbf{lvt2k-1sent} and \textbf{lvt5k-1sent} \cite{nallapati2016abstractive} utilize a trick to control the vocabulary size to improve the training efficiency.
	%\item \textbf{Read-Again} \cite{zeng2016efficient}
	\textbf{SEASS} \cite{zhou2017selective} integrates a selective gated network into the seq2seq framework to control the information flow from encoder to decoder.
	\textbf{DRGD} \cite{li2017deep} proposes a deep recurrent generative decoder to enhance the modeling ability of latent structures in the target summaries.
	%\textbf{GBN} \cite{chen2018generative} proposes a generative bridging network in which a bridge module is introduced to assist the training of the sequence prediction model.
	
	\subsection{Experimental Settings}
	For the experiments on the English dataset of Gigawords, we set the dimension of word embeddings to 300, and the dimension of hidden states and latent variables to 500.
	The maximum length of documents and summaries is 100 and 50 respectively.
	For DUC-2004, the maximum length of summaries is 75 bytes.
	For the dataset of LCSTS, the dimension of word embeddings is 350.
	We also set the dimension of hidden states and latent variables to 500.
	The maximum length of documents and summaries is 120 and 25 Chinese characters respectively. The damping factor $d$ of the PageRank algorithm for the unsupervised attention learning is set to 0.9.  
	The beam size of the decoder is set to 10.
	Our neural network based framework is implemented using Theano \cite{2016arXiv160502688short}.% on a single Tesla K80 GPU.

	\section{Results and Discussions}
	
	\subsection{ROUGE Evaluation}
	\label{sec:rouge}

	\begin{table}[!t]
		\centering
		\begin{tabular}{p{3cm} c c c}
			\hline
			\textbf{System}  & \textbf{R-1} & \textbf{R-2} & \textbf{R-L}  \\
			\hline
			ABS       & 29.55 & 11.32 & 26.42  \\
			ABS+       & 29.78 & 11.89 & 26.97  \\
			RAS-LSTM       & 32.55 & 14.70 & 30.03  \\
			RAS-Elman       & 33.78 & 15.97 & 31.15  \\ 
			ASC-FSC$_1$       & 34.17 & 15.94 & 31.92  \\
			%RNN-MRT     & 36.54 & 16.59 & 33.44  \\
			lvt2k-1sent     & 32.67 & 15.59 & 30.64  \\
			lvt5k-1sent     & 35.30 & 16.64 & 32.62  \\
			%\hline
			%GBN    & 35.26 & 17.22 & 32.67  \\
			SEASS     & 36.15 & 17.54 & 33.63  \\
			\hline
			seq2seq (our version)       & 34.49 & 16.79 & 33.06 \\
			seq2seq+SuAtt       & 35.80 & 17.02 & 33.25 \\
			seq2seq+UnAtt       & 35.91 & 17.16 & 33.38 \\
			\textbf{seq2seq+MAL}       & \textbf{36.39} & \textbf{17.37} & \textbf{33.82} \\
			\hline
			DRGD       & 36.27 &17.57 & 33.62  \\
			DRGD+MAL      & \textbf{36.30} & \textbf{17.77} & \textbf{33.64}  \\
			\hline
		\end{tabular}
		\caption{ROUGE-F1 on Gigawords.}
		\label{tab:rouge-agiga}

		\vspace{0mm}
	\end{table}

	\begin{table}[!t]
		\centering
		\begin{tabular}{p{3cm} c c c}
			\hline
			\textbf{System}  & \textbf{R-1} & \textbf{R-2} & \textbf{R-L} \\
			\hline
			TOPIARY & 25.12 & 6.46 & 20.12  \\
			MOSES+ & 26.50 & 8.13 & 22.85  \\
			ABS       & 26.55 &	7.06 &	22.05  \\
			ABS+       & 28.18 & 8.49 & 23.81  \\
			RAS-Elman       & 28.97 & 8.26 & 24.06  \\
			RAS-LSTM       & 27.41 & 7.69 & 23.06  \\
			LenEmb       & 26.73 & 8.39 & 23.88  \\
			lvt2k-1sen       & 28.35 & 9.46 & 24.59  \\
			lvt5k-1sen & 28.61 & 9.42 & 25.24  \\
			%Read-Again       & 29.74 & 9.44 & 25.94 \\
			SEASS       & 29.21 & 9.56 & 25.51 \\
			%DRGD       & 28.99 & 9.72 & 25.28 \\
			\hline
			seq2seq (our version)       & 28.82 & 9.47 & 25.27 \\
			seq2seq+SuAtt       & 29.46 & 9.92 & 25.85 \\
			seq2seq+UnAtt       & 29.13 & 9.96 & 25.63 \\
			\textbf{seq2seq+MAL}       & \textbf{29.49} & \textbf{10.13} & \textbf{25.91} \\
			\hline
			DRGD       & 28.99 & 9.72 & 25.28 \\
			DRGD+MAL      & \textbf{29.04} & \textbf{9.90} & \textbf{25.31} \\
			\hline
		\end{tabular}
		\caption{ROUGE-Recall on DUC-2004.}
		\label{tab:rouge-duc04}

		\vspace{0mm}
	\end{table}

	\begin{table}[!t]
		\centering
		\begin{tabular}{p{3cm} c c c}
			\hline
			\textbf{System}  & \textbf{R-1} & \textbf{R-2} & \textbf{R-L} \\
			\hline
			RNN       & 21.50 & 8.90 & 18.60  \\
			RNN-context       & 29.90 & 17.40 & 27.20  \\
			CopyNet       & 34.40 & 21.60 & 31.30  \\
			RNN-distract & 35.20 & 22.60 & 32.50  \\
			DRGD       & 36.99 & 24.15 & 34.21 \\
			\hline
			seq2seq (our version)       & 35.38 & 23.00 & 32.85 \\
			seq2seq+SuAtt       & 36.80 & 24.20 & 34.45 \\
			seq2seq+UnAtt       & 37.00 & 24.20 & 34.39 \\
			\textbf{seq2seq+MAL}       & \textbf{37.04} & \textbf{24.65} & \textbf{34.70} \\
			\hline
			DRGD       & 36.99 & 24.15 & 34.21 \\
			DRGD+MAL      & \textbf{37.36}  & \textbf{24.73}  & \textbf{34.65}  \\
			\hline
		\end{tabular}
		\caption{ROUGE-F1 on LCSTS.}
		\label{tab:rouge-lcsts}

		\vspace{0mm}
	\end{table}
	
	The results on the English datasets of Gigawords and DUC-2004 are shown in Table~\ref{tab:rouge-agiga} and Table~\ref{tab:rouge-duc04} respectively. 
	Among the ablations, ``\textbf{seq2seq (our version)}'' is the typical attention based seq2seq framework implemented by us. ``\textbf{seq2seq+SuAtt}'' is the ablation method only considering supervised attention information.
	``\textbf{seq2seq+UnAtt}'' only considers unsupervised attention information.
	``\textbf{seq2seq+MAL}'' is our proposed framework.
	%To further illustrate the potential superiority of our MAL framework, we integrate it with one strong baseline method DRGD \cite{li2017deep} and obtain a new approach: ``\textbf{DRGD+MAL}''. 
	From the experimental results, we can see that our MAL framework performs better than the typical seq2seq method as well as some other strong comparisons, which means that the multi-attention context information can indeed improve the performance of the typical seq2seq summarization models.
	It is worth noting that the methods lvt2k-1sent and lvt5k-1sent utilize linguistic features such as parts-of-speech tags, named-entity tags, and TF and IDF statistics of the words as part of the document representation.
	Generally, more useful features can indeed improve the performance. Nevertheless, our framework is still better than them which demonstrates the effectiveness of our salience detection components. 
	
	The results on the Chinese dataset LCSTS are shown in Table~\ref{tab:rouge-lcsts}.
	Our MAL also achieves the best performance. Although CopyNet employs a copying mechanism to improve the summary quality, RNN-distract considers attention information diversity in their decoders, and DRGD integrates a recurrent variational auto-encoder into the typical seq2seq framework, our model is still better than these methods demonstrating that the effectiveness of the incorporation of the multi-attention context information.
	It is expectable that integrating the copying mechanism and coverage diversity in our framework will further improve the summarization performance. 
	
	\subsubsection{Highlight Discussion}
	
	Note that in our framework, we integrate the multi-attention information with a simple base model, namely, the attention based seq2seq model. Thus the performance of the whole framework is indeed limited. And the evaluation results are not as good as some very strong recent methods, such as SEASS \cite{zhou2017selective}, Pointer-Generator \cite{see2017get}, and the Reinforced model \cite{paulus2017deep}. However, the purpose of this work is to investigate the performance of applying the traditional salience detection intuitions in the simple attention based seq2seq framework, and such a simple base model allows the conclusions not biased by other modeling complications. The experimental analysis can demonstrate its effectiveness, therefore, our study in this paper not only reminds the peer researchers that the crucial salience detection component for summarization should be reexamine in the scope of neural network based models, but also presents a practical approach to solving this problem. If the two types of attention signals are appropriately integrated into the above recent models, we believe that their performance can be improved as well. Moreover, our attention learning framework can also help revise the design of the copy mechanism as well as the coverage modeling strategy. All these are worthwhile directions to investigate for the future works.

	\subsection{Attention Analysis}
	
	\begin{table}[!t]
		\centering
		\begin{tabular}{p{2.6cm} c c c}
			\hline
			\textbf{System}  & \textbf{Giga} & \textbf{DUC} & \textbf{LCSTS} \\
			\hline
			SuAtt       & 30.97 &  29.14 & 24.97  \\
			UnAtt       & 21.38 &  20.16 & 17.87 \\
			\hline
		\end{tabular}
		\caption{ROUGE-1 evaluation for the top-10 words extracted from SuAtt and UnAtt.}
		\label{tab:rouge-attn}

		\vspace{0mm}
	\end{table}
	
	\begin{table}[!t]
		\centering
		\begin{tabular}{p{7.2cm}}
			\hline
			\hline
			\textbf{S(1)}: japan 's toyota team europe were banned from the world rally championship for one year here on friday in a crushing ruling by the world council of the international automobile federation fia.\\
			\textbf{Golden}: toyota are banned for a year.\\
			\textbf{SuAtt}: toyota, rally, world, europe, banned, championship, team, ruling, year, fia\\
			\textbf{UnAtt}: world, council, europe, federation, international, japan, ruling, friday, championship, banned\\
			\hline
			\textbf{S(2)}: a powerful bomb exploded outside a navy base near the sri lankan capital colombo tuesday, seriously wounding at least one person, military officials said.\\
			\textbf{Golden}: bomb attack outside srilanka navy base.\\
			\textbf{SuAtt}: sri, bomb, base, navy, colombo,  ankan, powerful, military, wounding, exploded\\
			\textbf{UnAtt}: sri, military, capital, tuesday, bomb, powerful, navy, exploded, base, officials\\
			\hline
			\textbf{S(3)}: palestinian prime minister ismail haniya insisted friday that his hamas-led government was continuing efforts to secure the release of an israeli soldier captured by      militants.\\
			\textbf{Golden}: efforts still underway to secure soldier's release: hamas pm.\\
			\textbf{SuAtt}: palestinian, release, haniya, hamas-led, soldier, israeli, government, secure, efforts, minister\\
			\textbf{UnAtt}: government, prime, palestinian, friday, israeli, militants,	efforts, continuing, minister, secure\\
			\hline
			\hline
		\end{tabular}
		\caption{Top-10 words extracted from SuAtt and UnAtt receptively for samples in Gigawords.}
		\label{tab:topk}
		\vspace{0mm}
	\end{table}
	
	We regard the supervised attention and the unsupervised attention as the salience score for the words in the source text. So we also design experiments to verify the performance of the two attention mechanisms for finding important words.
	For each input sequence, $\mathbf{a}^s$ and $\mathbf{a}^u$ are the two attention vectors obtained by supervised attention learning and unsupervised attention learning respectively. The element value $a_i \in \mathbf{a}$ represents the word salience score. Therefore, we can select the top-$k$ words from the input sequence according to the salience scores in $\mathbf{a}^s$ and $\mathbf{a}^u$.
	Intuitively, the extracted top-$k$ words are very important and may have a large overlapping with the ground truth summary. To verify it quantitatively, we regard the top words as summaries and conduct ROUGE evaluation on them. Because the order of the top words is ignored, so we employ the F-measure score of  ROUGE-1 as the evaluation metric. The experimental results on those three datasets are given in Table~\ref{tab:rouge-attn}. We set $k$ to 10 here. The results illustrate that both methods can extract the important words from the source text, and the quality of the top words extracted from the supervised attention $\mathbf{a}^s$, i.e., SuAtt, is better than those extracted from the unsupervised attention $\mathbf{a}^u$, i.e. UnAtt. This adheres to our intuition because the SuAtt method can obtain stronger supervision signals than the unsupervised method UnAtt. 
	
	However, from the ROUGE results presented in Tables~\ref{tab:rouge-agiga},  \ref{tab:rouge-duc04}, and  \ref{tab:rouge-lcsts}, we find that the performance of seq2seq+UnAtt is similar to or even better than seq2seq+SuAtt. This phenomenon may be because that both the method of seq2seq and SuAtt can receive supervision signals to guide the training, but UnAtt is an unsupervised salience detection method which may find some complementary information to further improve the summarization performance.  
	In order to show the differences vividly, we present the extracted top words in Table~\ref{tab:topk}. And all the words are ranked based on the corresponding salience scores. From the results we know that SuAtt and UnAtt can indeed assign large salience scores to the important words. For instance, SuAtt can extract words of ``toyoda'', ``banned'', and ``year'' which are the core elements of the golden summary ``toyota are banned for a year''. The result of UnAtt is more diversified. Although the performance of SuAtt and UnAtt are different, the integration of them performs well in the quantitative evaluation experiments in the previous subsection, which may be because that different attention methods can capture different aspects of the source text and they can complement each other.

	\subsection{Summary Case Analysis}
	
	\begin{table}[!t]
		\centering
		\begin{tabular}{p{7.2cm}}
			\hline
			\hline
			\textbf{S(1)}: japan 's toyota team europe were banned from the world rally championship for one year here on friday in a crushing ruling by the world council of the international automobile federation fia.\\
			\textbf{Golden}: toyota are banned for a year.\\
			\textbf{seq2seq}: toyota 's world rally europe banned from world rally championship.\\
			\textbf{MAL}: \textbf{toyota barred from world rally championship.}\\
			\hline
			\textbf{S(2)}: slovaks started voting at \#:\#\# am on saturday in elections to the \#-seat parliament, with centre-right prime minister mikulas dzurinda fighting to continue far-reaching but painful reforms.\\
			\textbf{Golden}: slovaks start voting in legislative elections. \\
			\textbf{seq2seq}: slovakia's parliament begins voting.\\
			\textbf{MAL}: \textbf{slovaks start voting in early elections.}\\
			\hline
			\textbf{S(3)}: the thai government has set aside \#\#\# million baht about \#\#.\#\# million u.s. dollars to support new eco-tourism plans during \#\#\#\#-\#\#\#\# , according to a report of the thai news agency tna tuesday.\\
			\textbf{Golden}: thai government to support eco-tourism.\\
			\textbf{seq2seq}: thailand to support new eco-tourism in \#\#\#\#-\#\#\#\#.\\
			\textbf{MAL}: \textbf{thailand to support new eco-tourism plans.}\\
			\hline
			\hline
		\end{tabular}
				\caption{Examples of the generated summaries.}
		\label{tab:cases}

	\end{table}
	
	Finally, some examples of the source texts, golden summaries, and the generated summaries by the typical attention-based seq2seq framework and our proposed MAL framework are shown in Table~\ref{tab:cases}. From these cases we can see that the generated summaries by MAL generally have better quality. Moreover, because of the attention learning components for salience detection, our framework has the ability to assign small salience scores to unimportant words and uninformative symbols, while the summary generated by seq2seq contains more noisy symbols as the cases shown in S(3). 
	%Our framework is able to avoid to generate such noise.

	\section{Related Works}
	\label{sec:relatedworks}
	
	Automatic summarization is the process of automatically generating a summary that retains the most important content of the original text document \cite{nenkova2012survey}.
	Conventional summarization methods can be classified into three categories: extraction-based methods \cite{erkan2004lexrank,min2012exploiting}, compression-based methods \cite{li2013document,wang2013sentence,li2017cascaded}, and abstraction-based methods \cite{barzilay2005sentence,lidong15absmds}. \footnote{Some researchers regard the compression approach as a special case of the extraction approach.} 
	%However, most of the traditional models employ indirect approaches to generate new sentences. Foe instance, \citet{barzilay2005sentence} employed sentence fusion to generate a new sentence. \cite{lidong15absmds} proposed a more fine-grained fusion framework, where new sentences are generated by selecting and merging salient phrases.
	
	Recently, some researchers employ neural network based frameworks to tackle the abstractive summarization problem and obtain encouraging performance.
	\citet{rush2015neural} proposed a neural model with local attention modeling, which is trained on the Gigaword corpus, but combined with an additional log-linear extractive summarization model with handcrafted features.
	\citet{nallapati2016abstractive} utilized a trick to control the vocabulary size to improve the training efficiency.
	\citet{gu2016incorporating} integrated a copying mechanism into a seq2seq framework to improve the quality of the generated summaries.
	\citet{chen2016distraction} proposed a new attention mechanism that not only considers the important source segments, but also distracts them in the decoding step in order to better grasp the overall meaning of input documents.
	\citet{miao2016language} extended the seq2seq framework and proposed a generative model to capture the latent summary information.
	\citet{zhou2017selective} integrated a selective gated network into the seq2seq framework to control the information flow from encoder to decoder.
	\cite{li2017deep} proposed a deep recurrent generative decoder to enhance the modeling ability of latent structures in the target summaries.
	\cite{see2017get} employed pointer networks and converge mechanism to improve the quality of the generated summaries.
	\citet{paulus2017deep} proposed a reinforcement learning based framework to enhance the performance of summarization.
	\citet{chen2018generative} proposes a generative bridging network in which a bridge module is introduced to assist the training of the sequence prediction model.
	\citet{li2018actor} employ actor-critic training paradigm to enhance the quality of the generated summaries.
	
	Meanwhile, some researchers also combine the traditional salience estimation methods into the seq2seq frameworks in order to enhance the summarization performance. \citet{tan2017abstractive} incorporated the graph-based attention information obtained by the PageRank algorithm into their framework. \citet{hsu2018unified} weighted the attention mechanism using sentence salience information calculated by a traditional supervised method.
	In contrast, we consider both the supervised salience information and unsupervised salience information in our framework to generate better summaries.
	
	\section{Conclusions}
	
	In this work, we investigate the effect of adding the traditional salience detection of text summarization back to the typical attention-based seq2seq framework for abstractive summarization. We propose a Multi-Attention Learning (MAL) framework which contains two new attention learning components, namely, supervised attention learning and unsupervised attention learning, for salience estimation.
	The salience information obtained based on these two types of attentions is incorporated with the typical attention mechanism in the decoder to conduct the summary generation. 
	Extensive experiments on some benchmark datasets in different languages demonstrate the effectiveness of the proposed framework for the task of abstractive summarization.

\bibliography{acl2020}
\bibliographystyle{acl_natbib}

\end{document}